\documentclass[letterpaper,10pt,journal,twoside]{IEEEtran}

\usepackage[utf8]{inputenc}
\usepackage[T1]{fontenc}

\usepackage{url}
\usepackage{booktabs}
\usepackage{amsfonts}
\usepackage{amsmath}
\usepackage{nicefrac}
\usepackage{microtype}
\usepackage{xcolor}
\usepackage{graphicx}
\usepackage{float}
\usepackage{multirow}

\usepackage[numbers,sort&compress]{natbib}

\usepackage{capt-of}

\usepackage{dblfloatfix}
\usepackage{placeins}

\usepackage{hyperref}
\usepackage{etoolbox}
\newcommand{\authorlink}[2]{{\hypersetup{hidelinks}\href{#1}{#2}}}

\makeatletter
\patchcmd{\@makecaption}{\ifCLASSOPTIONconference}{\iftrue}{}%
  {\PackageError{arxiv}{Could not enable centered short captions}{Check the IEEEtran caption definition.}}
\makeatother

\newcommand{\ours}{TartanMatch}
\newcommand{\na}{--}

\begin{document}

\title{
TartanMatch: Towards Universal Dense Correspondence Across Modalities\\[-0.3em]
{\large\sffamily\bfseries\hypersetup{hidelinks}%
\href{https://tartanmatch.github.io/}{\textcolor{blue!60!black}{tartanmatch.github.io}}}
}

\author{%
\authorlink{https://www.linkedin.com/in/hyeokjoon-kwon/?isSelfProfile=false}{Hyeokjoon Kwon}$^{*1}$,
\authorlink{https://caijiting.github.io/}{Jiting Cai}$^{*\dagger 1}$,
\authorlink{https://www.linkedin.com/in/ruogu-li-47a896364/}{Ruogu Li}$^{*1}$,
\authorlink{https://www.linkedin.com/in/kritanbhandari/}{Kritan Bhandari}$^{2}$,
\authorlink{https://geethikah21.github.io/}{Geethika Hemkumar}$^{1}$,\\
\authorlink{https://www.parvmaheshwari.com/}{Parv Maheshwari}$^{1}$,
\authorlink{http://yuhengqiu.com/}{Yuheng Qiu}$^{1}$,
\authorlink{https://infinity1096.github.io/}{Yuchen Zhang}$^{1}$,
\authorlink{https://www.ri.cmu.edu/ri-faculty/sebastian-scherer/}{Sebastian Scherer}$^{1}$, and
\authorlink{http://www.wangwenshan.com/}{Wenshan Wang}$^{1}$%
\thanks{$^{*}$Equal contribution; the listing order is random.}%
\thanks{$^{\dagger}$Corresponding author.}%
\thanks{$^{1}$The authors are with the Robotics Institute, Carnegie Mellon University, Pittsburgh, PA, USA
(e-mail: \texttt{\{joonk2, jitingc, ruoguli, ghemkuma, parvm, yuhengq, yuchenz7, basti, wenshanw\}@andrew.cmu.edu}).}%
\thanks{$^{2}$Kritan Bhandari is with Fisk University, Nashville, TN, USA (e-mail: \texttt{kbhandari07@my.fisk.edu}). His contribution to this work was completed during the Robotics Institute Summer Scholars (RISS) program at Carnegie Mellon University.}%
}
\IEEEaftertitletext{%
\vspace{-4.0em}
\begin{center}
\includegraphics[width=1\textwidth]{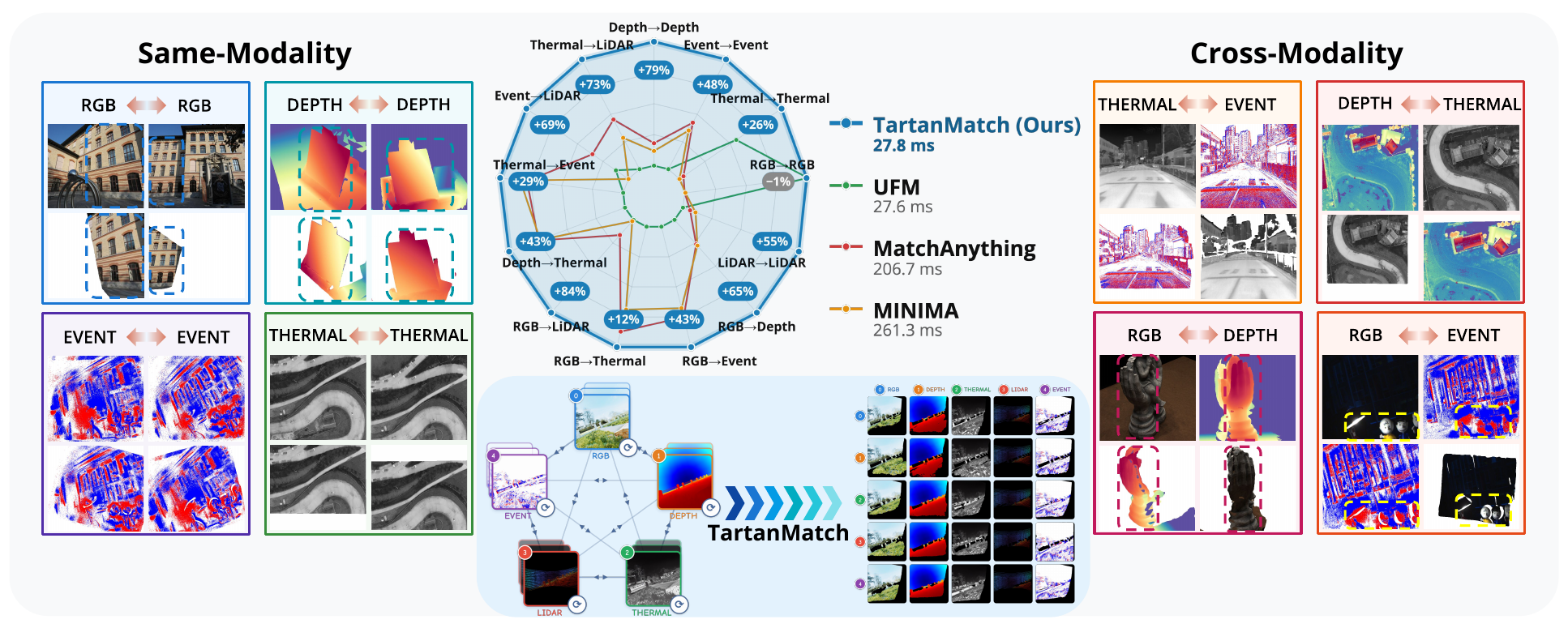}
\vspace{-0.8em}
\captionof{figure}{%
\textbf{TartanMatch}: a unified model for dense correspondence
across five sensing modalities---RGB, event, thermal, depth, and
LiDAR. Given any pair of two modalities (25
combinations), TartanMatch estimates dense flow, visualized by warping
each target onto each source. The radar chart compares endpoint error
(EPE) against multi-modal correspondence baselines, averaged over all
benchmarks per modality pair; for visualization, each axis shows $\log(1/\mathrm{EPE})$,
min--max normalized per axis, so a larger polygon indicates lower error. Annotations show relative EPE improvement over the best competing multi-modal method per pair. Finetuned from UFM, TartanMatch
improves every other modality pair while preserving RGB$\to$RGB
accuracy ($<$1\%).
}
\label{fig:teaser}
\end{center}
}
\maketitle

\begin{abstract}
Robotic platforms combine multi-modal sensors to operate
across diverse conditions, but most dense correspondence methods
are limited to RGB-RGB or a small set of modality pairs.
Training a model for each modality pair scales poorly with the number of sensors
and forgoes potential positive transfer across modality pairs.
We introduce TartanMatch, a unified model for dense correspondence
across all 25 ordered pairs among RGB, event, thermal, depth, and
projected LiDAR. TartanMatch adapts an RGB-pretrained correspondence
model to all modality pairs within
one shared network using lightweight modality adaptation and a mixture of real and synthetic data.
On zero-shot real-world benchmarks, TartanMatch reduces Endpoint Error (EPE)
over the strongest baselines by 61.9\% for cross-modal matching
and 49.6\% for same-modal matching. We further show that joint training across all modality pairs
improves average matching accuracy over pair-specific training,
especially on challenging cross-modal pairs. These results demonstrate that a single model achieves broad multi-modal coverage and state-of-the-art matching accuracy.
\end{abstract}

\section{Introduction}
\label{sec:introduction}

Dense correspondence---estimating where each pixel in one observation appears in another---is a core primitive for robotic perception tasks such as visual odometry, place recognition, and 3D reconstruction.
Although correspondence estimation has advanced rapidly for RGB images, robots do not perceive the world through RGB alone.
Modern robotic platforms combine multiple sensors with distinct yet complementary sensing capabilities: event cameras capture rapid motion over a wide dynamic range, thermal cameras remain informative in darkness, smoke, and adverse weather, and depth or LiDAR sensors provide geometry when appearance is ambiguous~\citep{zhu2018mvsec,gehrig2021dsec,dhrafani2025firestereo,cho2026dsert}.
Robust robotic perception therefore calls for correspondence between any sensors informative in the environment.

Existing correspondence methods cover only parts of this multi-modal sensor space.
Sensor-specific methods are designed for a same-modal pair setting: RGB methods span optical flow~\citep{teed2020raft,wang2024searaft} and wide-baseline matching~\citep{sun2021loftr,edstedt2024roma}, with recent models unifying both settings~\citep{zhang2026ufm}, yet operate only on RGB--RGB pairs; likewise, event-based methods match only Event--Event pairs~\citep{gehrig2021eraft,zhang2024ematch}. Cross-modal matchers extend matching to pairs such as RGB--Depth or RGB--Thermal, but each typically supports only one such modality pair~\citep{tuzcuoglu2024xoftr,tong2025sc3ef,shrivastava2025cmrw}.
Recent methods broaden coverage to several cross-modal pairs by fine-tuning RGB matchers on synthetic pairs that combine an RGB view with a non-RGB view generated from RGB imagery~\citep{ren2025minima,he2025matchanything}. However, this training remains RGB-anchored---the model never sees a pair in which both views are non-RGB---and represents non-RGB inputs using fixed RGB-like renderings. Such renderings can suppress sensor-specific structure useful for accurate correspondence, such as event timing and depth validity. Together, these limitations leave a central question: \textit{can one unified model accurately estimate dense correspondence for every ordered pairing among diverse sensor modalities?}

This unification may also improve matching accuracy through positive transfer across modality pairs. Although different modality sensors measure distinct physical quantities, every modality pair poses the same matching task: identifying pixels in two views that correspond to the same physical point. This common task motivates a second question: \textit{does joint training across all modality pairs improve matching accuracy over pair-specific training?}

Motivated by these questions, we introduce TartanMatch, a unified model for any-pair multi-modal dense correspondence among RGB, event, thermal, depth, and projected LiDAR. Given source and target observations drawn from these five modalities, TartanMatch predicts dense flow and covisibility. A single set of weights thus covers all 25 modality pairs, replacing a collection of pair-specific models.

TartanMatch adapts UFM~\citep{zhang2026ufm}, an RGB-pretrained correspondence model, to support dense matching across five modalities and all 25 source--target pairs. Rather than converting non-RGB modality into fixed RGB-like rendering, we represent each modality in a form that retains its own measurement structure and, where applicable, use lightweight learned projections to map this representation to UFM's three-channel input. These inputs are then processed by the same UFM backbone, enabling one shared model to handle every modality pair. To effectively adapt the pretrained model to these new modalities, we train TartanMatch in two stages: we first freeze the pretrained encoder of UFM while optimizing the input projections and downstream layers, then fine-tune the full network end to end. We empirically observed that this schedule improves performance over single-stage training.

Training a unified matcher is bottlenecked by data: no single real-world dataset covers all five modalities with correspondence or flow labels. We therefore assemble a training suite that combines synthetic sources with real indoor, outdoor, aerial, and driving scene captured from diverse datasets, rather than solely relying on synthetic modality generation as in prior works~\cite{ren2025minima, he2025matchanything, tuzcuoglu2024xoftr}. To construct supervision, we use provided flow labels when available and otherwise derive flow from depth and camera poses, refining inaccurately derived flow with off-the-shelf matchers~\citep{edstedt2025romav2,kroeger2016dis}. Separately, we synthesize thermal views to expand coverage of this underrepresented modality~\citep{lee2026thera}.

On zero-shot real-world benchmarks covering a diverse set of same-modal and cross-modal pairs across all five modalities, TartanMatch reduces average EPE relative to the strongest baseline for each benchmarked modality pair by 61.9\% on cross-modal matching and 49.6\% on same-modal matching.  Moreover, we show that joint multi-modal training reduces average EPE relative to pair-specific training by 34.1\% across cross-modal pairs and 17.8\% across same-modal pairs, improving 11 of the 15 evaluated pairs, with the largest gains on challenging cross-modal pairs.

In conclusion, our contributions are:
\begin{itemize}
  \item \textbf{Any-pair multi-modal dense correspondence.} We introduce a unified model that predicts dense flow and covisibility for every ordered pair among RGB, event, thermal, depth, and projected LiDAR. We further show that joint training across all modality pairs improves average per-pair performance relative to training a separate model on each pair in isolation.

  \item \textbf{Lightweight modality adaptation.} We adapt an RGB-pretrained correspondence model using modality-specific representations and lightweight learned projections, which outperform the fixed RGB-like rendering of prior works. A two-stage training schedule, freezing the encoder before end-to-end fine-tuning, further aids adaptation and improves matching accuracy.

  \item \textbf{Multi-modal training and evaluation suite.} We build a training data mixture that combines real and synthetic sources across all 25 modality pairs, spanning indoor, outdoor, aerial, and driving scenes. The benchmark evaluation suite covers all five modalities on real-world zero-shot datasets.
\end{itemize}

\section{Related Work}
\label{sec:related_work}

\subsection{Same-Modal Correspondence}

Same-modal correspondence matches inputs from a single sensor modality. Most work here concentrates heavily on RGB--RGB, and methods for other modalities such as event and thermal are developed independently rather than as instances of a shared model. Within this setting, optical-flow methods estimate dense motion
between temporally adjacent
frames~\citep{teed2020raft,wang2024searaft}, whereas wide-baseline matching
methods target larger viewpoint and appearance
changes~\citep{sun2021loftr,edstedt2024roma}. Recent unified RGB models, such as UFM, bridge these two regimes within a
shared architecture~\citep{zhang2026ufm}. Beyond RGB, current methods still remain sensor-specific such
 as E-RAFT and EMatch for event
streams~\citep{gehrig2021eraft,zhang2024ematch} and SuperThermal for thermal
images~\citep{lu2021superthermal}. TartanMatch extends dense correspondence across modalities, treating same-modal matching as a subset of an any-pair correspondence problem spanning RGB, event, thermal, depth, and projected LiDAR.

\subsection{Cross-Modal Correspondence}
Cross-modal correspondence estimates matches between inputs from different
sensor modalities, but most existing methods are limited to a few modality pairs. XoFTR and SC3EF focus on RGB--Thermal
matching~\citep{tuzcuoglu2024xoftr,tong2025sc3ef}, while Walters et al.~\citep{walters2021multispectral} estimates dense multispectral correspondence for RGB--FIR and
evaluates transfer to RGB--NIR.
MINIMA and MatchAnything broaden this coverage by fine-tuning existing RGB matchers on pairs whose non-RGB view is generated from a real RGB image: MINIMA uses a diffusion model for IR and Depth Anything V2 for depth~\citep{ren2025minima}, and MatchAnything uses image-translation models for thermal and nighttime views~\citep{he2025matchanything}. Correspondence labels are therefore inherited from the RGB source rather than measured, and no training pair contains two non-RGB views.
Moreover, their
standard image backbones encode non-RGB inputs through fixed RGB-like renderings, which
can collapse sensor-specific information such as event timing or geometric depth.
TartanMatch instead trains one model on all 25 directed pairs among five modalities, including pairs in which neither view is RGB. Each modality is encoded in its own representation and mapped to the shared backbone through lightweight learned projections.

\subsection{Multi-modal Data for Supervision}
Prior work has introduced a rich collection of multi-modal sensory datasets, from synthetic engines that render several modalities in perfect alignment~\citep{wang2020tartanair} to real captures spanning RGB, event, thermal, and depth or LiDAR sensors~\citep{gehrig2021dsec,zhu2018mvsec,cho2026dsert,dhrafani2025firestereo,liu2023mtv}.
However, these resources are distributed across datasets with different sensor subsets and rarely provide cross-modal correspondence labels directly. Flow derived from depth and camera geometry may also be
inaccurate because of invalid measurements or calibration errors. Constructing a broad correspondence training set from these resources therefore requires substantial data integration and label processing.
Consequently, prior methods have
typically relied on RGB-derived synthetic modality pairs~\citep{ren2025minima,he2025matchanything} or individual
sensor datasets with native supervision~\citep{gehrig2021eraft,tuzcuoglu2024xoftr}.
To address this limitation, we integrate
diverse real and synthetic datasets into a comprehensive training and evaluation suite,
deriving flow from geometry and refining unreliable
labels.

\section{Method}
\label{sec:method}

\begin{figure*}[t!]
  \centering
  \includegraphics[width=0.98\textwidth]{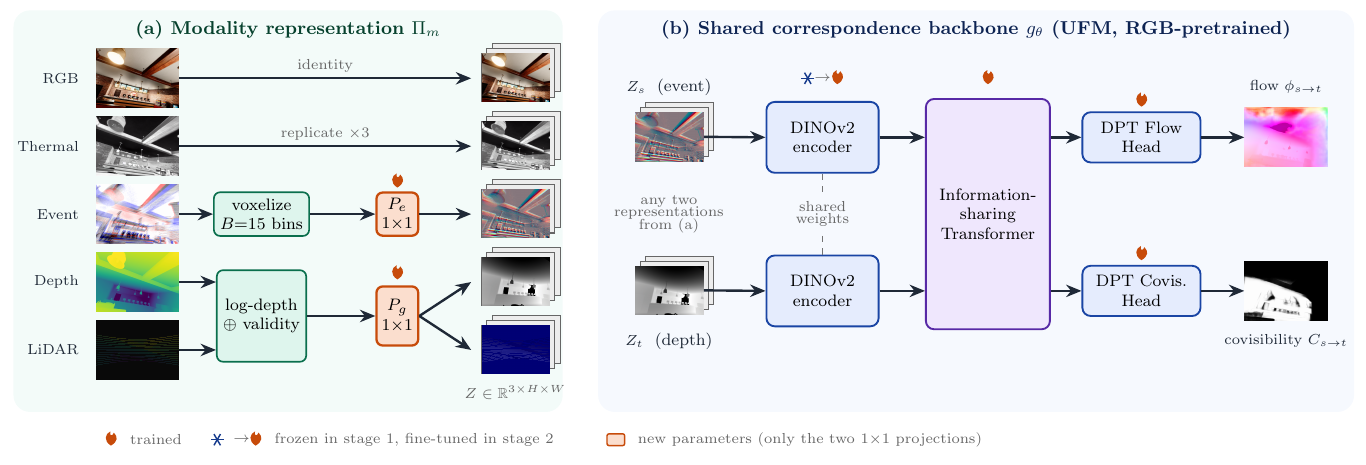}
  \caption{TartanMatch architecture.
  A modality-specific adaptation maps every sensor to the three-channel input the pretrained backbone expects: RGB and thermal enter directly, events are voxelized into $B{=}15$ temporal bins and mapped by a learned $1{\times}1$ convolution, and depth and projected LiDAR share a log-depth-plus-validity representation and a single learned $1{\times}1$ convolution.
  The UFM correspondence backbone (DINOv2 encoder, information-sharing transformer, DPT heads) is fully shared across all modality pairs.}
  \label{fig:architecture}
\end{figure*}

\subsection{Overview}
\label{sec:method_overview}
TartanMatch predicts dense correspondence between any pair drawn from five modalities: RGB, thermal, event, depth, and projected LiDAR, denoted $\mathcal{M}$.
Given a source input $X_s$ of modality $m_s\in\mathcal{M}$ and a target input $X_t$ of modality $m_t\in\mathcal{M}$, the model predicts
\begin{equation}
  \left(\phi_{s\rightarrow t},\, C_{s\rightarrow t}\right)
  = g_{\theta}\!\left(\Pi_{m_s}(X_s),\, \Pi_{m_t}(X_t)\right),
\end{equation}
where $\phi_{s\rightarrow t}\in\mathbb{R}^{2\times H\times W}$ is the source-to-target dense flow and
$C_{s\rightarrow t}\in[0,1]^{H\times W}$ is the per-pixel probability that a source pixel is covisible in the target view.
Each $\Pi_m$ is a modality-specific adaptation that maps a sensor's measurements to a three-channel tensor (Sec.~\ref{sec:modality_formatting}), and $g_\theta$ is a correspondence backbone shared by all modality pairs (Sec.~\ref{sec:shared_backbone}).

In this work, we instantiate $g_\theta$ with UFM~\citep{zhang2026ufm}, as
shown in Fig.~\ref{fig:architecture}. We retain UFM's DINOv2 encoder,
information-sharing transformer, and DPT prediction heads without
architectural modification. The pretrained backbone is adapted to the five modalities with a two-stage schedule on a multi-modal data mixture (Secs.~\ref{sec:training_procedure} and~\ref{sec:datasets}).

\subsection{Modality Adaptation}
\label{sec:modality_formatting}

The modality adapter $\Pi_m$ maps an input observation $X$ of modality $m$
to a three-channel tensor $Z=\Pi_m(X)\in\mathbb{R}^{3\times H\times W}$ that the pretrained backbone can consume.
The central design choice is how each sensor gets there: instead of hand-rendering non-RGB sensors as fixed RGB-like visualizations, as in prior cross-modal matching pipelines~\citep{ren2025minima,he2025matchanything}, we keep each modality in a representation suited to it and let a learned $1{\times}1$ projection decide how to expose it to the backbone.

\paragraph{RGB and Thermal}
RGB images are used as is, $Z_{\mathrm{rgb}}=X_{\mathrm{rgb}}$, and single-channel thermal intensity is replicated across the three channels.
Neither requires learned parameters.

\paragraph{Events}
Events are represented as a voxel volume~\citep{zhu2019voxelsevents}.
Given events
$\mathcal{E}=\{(x_i,y_i,t_i,p_i)\}_{i=1}^{N}$,
with $p_i\in\{-1,1\}$, recorded between two consecutive frames, we define
\begin{align}
t_i^{*}
&= (B-1)\frac{t_i-t_1}{t_N-t_1},\\[-0.3em]
V(x,y,t)
&= \sum_{i=1}^{N}
p_i\,k_b(x-x_i)\,k_b(y-y_i)\,k_b(t-t_i^{*}),
\end{align}
where $E\in\mathbb{R}^{H\times W\times B}$,
$x\in\{0,\ldots,W-1\}$,
$y\in\{0,\ldots,H-1\}$,
$t\in\{0,\ldots,B-1\}$, and
$k_b(a)=\max(0,1-|a|)$.
A single learned $1{\times}1$ projection $P_e:\mathbb{R}^{B}\rightarrow\mathbb{R}^{3}$ then produces $Z_{\mathrm{event}}=P_e(E)$, learning how to expose polarity and temporal structure to the pretrained three-channel backbone.

\paragraph{Depth and projected LiDAR}
We represent both depth and projected LiDAR using a depth map $d$ and a
validity mask $V\in\{0,1\}^{H\times W}$. For a depth observation, $d$ is
the provided depth image and $V$ identifies pixels with finite, valid
measurements; invalid pixels may arise from missing sensor returns,
out-of-range measurements, or infinite depth. A LiDAR observation is
originally a 3D point cloud. We transform its points into the camera
coordinate frame and project them onto the image plane, producing a sparse
depth map $d$ for which $V$ identifies pixels that receive a LiDAR return.
We refer to this rasterized representation as projected LiDAR.

Both modalities use the same geometric representation:
\begin{equation}
  G(d,V) =
  \begin{bmatrix}
    V\odot\log(1+d)\\
    V
  \end{bmatrix}
  \in\mathbb{R}^{2\times H\times W}.
\end{equation}
The first channel contains log-depth, with invalid values set to zero by
$V$, while the second explicitly preserves measurement validity. A shared
learned $1{\times}1$ projection
$P_g:\mathbb{R}^{2}\rightarrow\mathbb{R}^{3}$ produces
$Z_{\mathrm{geo}}=P_g(G(d,V))$ for both modalities.
\subsection{Shared Correspondence Backbone}
\label{sec:shared_backbone}

The three-channel tensors $Z_s=\Pi_{m_s}(X_s)$ and $Z_t=\Pi_{m_t}(X_t)$ are processed by the same UFM backbone regardless of the modality pair.
Both are encoded into patch tokens by the shared DINOv2 encoder; the two token sets are concatenated with view-index positional encodings and processed by the information-sharing transformer, which exchanges information between the two views; and DPT heads decode the source-view tokens into the flow field $\phi_{s\rightarrow t}$ and covisibility logits $\ell_C$, from which $C_{s\rightarrow t}=\sigma(\ell_C)$.
All backbone components are initialized from UFM's RGB-pretrained weights.
We initialize $P_e$ to pass the first three event bins through and $P_g$ to replicate log-depth across all three channels.
We adopt UFM's base configuration, which excludes its optional local refinement module.

Supervision also follows UFM.
Given ground-truth flow $\phi^\ast$ and covisibility mask $C^\ast$, we minimize
\begin{equation}
\begin{aligned}
  \mathcal{L}
  &=
  \frac{1}{\sum_p C^\ast(p)}
  \sum_p C^\ast(p)\,
  \rho_{\alpha,c}\!\left(
    \left\|\phi_{s\rightarrow t}(p)-\phi^\ast(p)\right\|_2
  \right)
  \\
  &\quad
  + \lambda\,\mathrm{BCE}(\ell_C, C^\ast),
\end{aligned}
\end{equation}
where $\rho_{\alpha,c}$ is the generalized Charbonnier loss. We use $\alpha=0.5$, $c=0.24$ and $\lambda=10$ following UFM.

\subsection{Two-Stage Training}
\label{sec:training_procedure}

We adapt the pretrained backbone to the multi-modal mixture in two stages.
Stage 1 freezes the encoder and trains the projections $P_e$ and $P_g$ together with the information-sharing transformer and DPT heads.
Stage 2 unfreezes the encoder and fine-tunes the full model end to end. The two stages address the mismatch between the new modalities and the encoder's pretrained input distribution: Stage 1 brings each modality closer to that distribution while the encoder is held fixed, whereas Stage 2 absorbs the residual shift that the projections cannot remove.
Sec.~\ref{sec:ablations} shows that this schedule improves over one-stage joint training.

\subsection{Training and Evaluation Data}
\label{sec:datasets}

TartanMatch extends UFM to multiple modalities and is trained on a multi-modal mixture combining synthetic imagery with real sensor data, summarized in Table~\ref{tab:datasets}.
From a dataset with \(k\) available modalities, all \(k^2\) ordered modality-pair types are eligible for training. We construct each 650k-sample epoch using fixed sampling rates at both the dataset and modality-pair levels. Table~\ref{tab:datasets} reports each dataset's aggregate share of the mixture, summed over its modality pairs.

\begin{figure}[tbp]
  \centering
  \includegraphics[page=1,width=\linewidth]{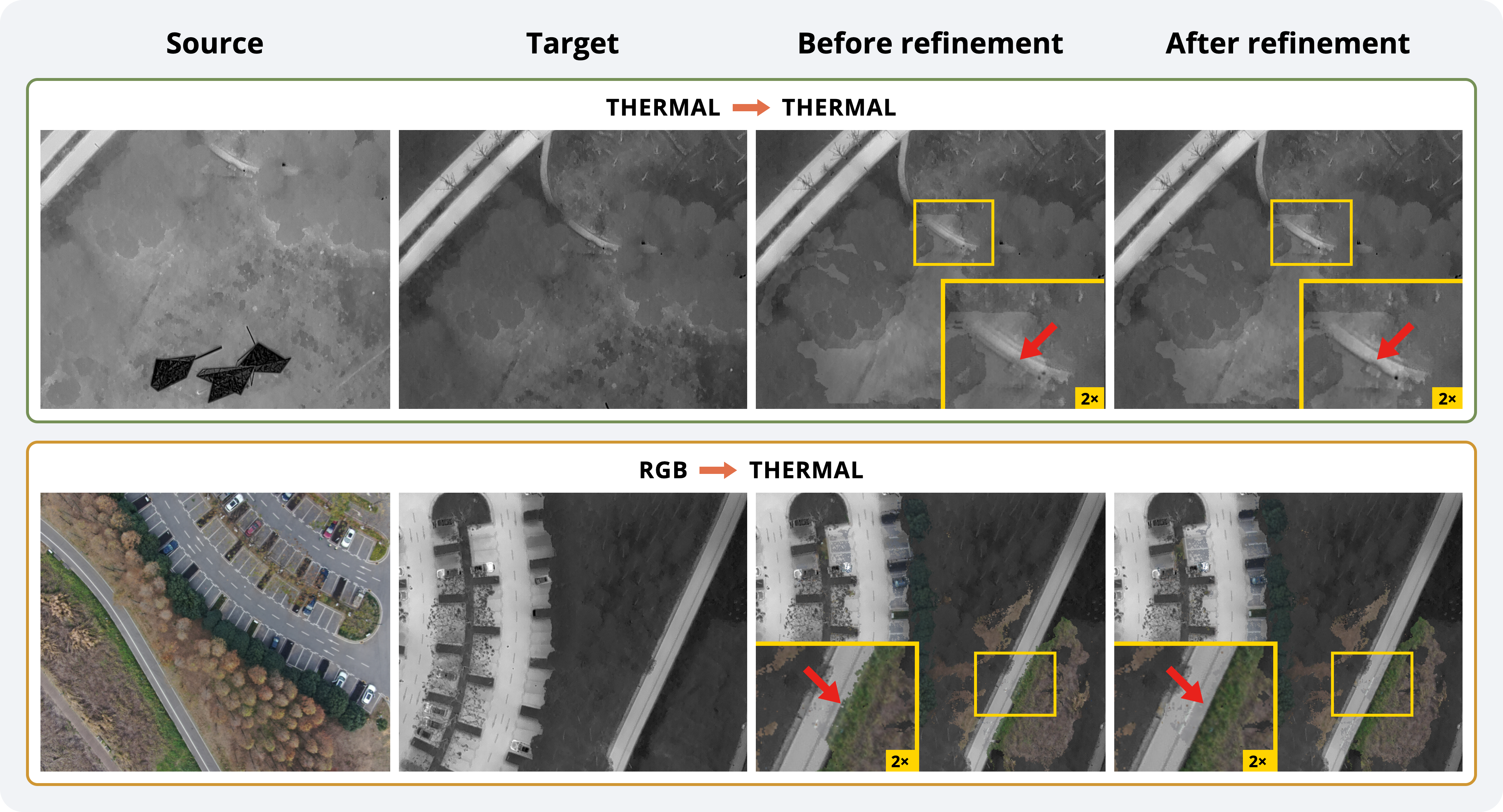}
  \caption{Ground-truth flow refinement on MTV.
  Each row warps the source onto the target using the flow label before and after refinement, for a thermal-to-thermal pair (top) and an RGB-to-thermal pair (bottom); insets magnify the outlined region.
  The misalignment left by the geometry-derived flow disappears after refinement.}
  \label{fig:gt_refinement}
\end{figure}

Ground-truth flow is either provided by the dataset or derived from depth, camera pose, and camera parameters.
On FIReStereo and MTV, where we empirically observed that the derived flow is sometimes inaccurate, we refine it: we warp the source view by the derived flow, estimate the residual displacement against the target with an off-the-shelf matcher (RoMa~v2~\citep{edstedt2025romav2} on FIReStereo; DIS~\citep{kroeger2016dis} on MTV), add it to the derived flow, and retain only forward--backward-consistent pixels, using thresholds of 1px for FIReStereo and 0.5px for MTV, respectively.
Fig.~\ref{fig:gt_refinement} shows the effect on MTV. Based on manual inspection of the refined labels, we retain FIReStereo's Hawkins 3 and 4 for training and MTV's race-track, country-road, and park scenes for evaluation.
Finally, to broaden thermal coverage, we synthesize thermal images with TherA~\citep{lee2026thera} for all MegaDepth training images and for 100k pairs from TartanAirV2 and TartanAirV2 WB.

\begin{table}[!tbp]
  \caption{Datasets in the TartanMatch training mixture and the zero-shot evaluation benchmarks.}
  \label{tab:datasets}
  \centering
  \footnotesize
  \resizebox{0.9\linewidth}{!}{%
  \begin{tabular}{lcccc}
    \toprule
    Dataset & OF/WB & Source & Modalities & Mix ratio \\
    \midrule
    \multicolumn{5}{l}{\textit{Training mixture}} \\
    \midrule
    TartanAirV2~\citep{wang2020tartanair} & OF & Synthetic & REDTL & 36.3\% \\
    MegaDepth~\citep{li2018megadepth} & WB & Both & RDT & 20.2\% \\
    DSEC~\citep{gehrig2021dsec} & OF & Real & RE & 10.9\% \\
    TartanAirV2 WB~\citep{wang2020tartanair, zhang2026ufm} & WB & Synthetic & RT & 9.7\% \\
    BlendedMVS~\cite{yao2020blendedmvs} & WB & Synthetic & RD & 4.3\% \\
    ScanNet++~\cite{yeshwanth2023scannet++} & WB & Real & RD & 4.3\% \\
    FlyingThings3D~\cite{mayer2016large} & OF & Synthetic & R & 3.7\% \\
    HyperSim~\cite{roberts2021hypersim} & WB & Synthetic & RD & 2.1\% \\
    FlyingChairs~\cite{dosovitskiy2015flownet} & OF & Synthetic & R & 1.8\% \\
    Spring~\cite{mehl2023spring} & OF & Synthetic & R & 1.8\% \\
    Kubric4D~\cite{greff2022kubric, van2024generative} & OF & Synthetic & RD & 1.6\% \\
    FIReStereo~\citep{dhrafani2025firestereo} & OF & Real & T & 1.5\% \\
    Habitat~\cite{szot2021habitat} & WB & Synthetic & RD & 0.5\% \\
    HD1K~\cite{kondermann2016hci} & OF & Real & R & 0.4\% \\
    Monkaa~\cite{mayer2016large} & OF & Synthetic & R & 0.4\% \\
    StaticThings3D~\cite{schroppel2022benchmark} & WB & Synthetic & RD & 0.4\% \\
    \midrule
    \multicolumn{5}{l}{\textit{Evaluation benchmarks}} \\
    \midrule
    MTV~\citep{liu2023mtv} & OF & Real & RDT & \na \\
    DSERT-RoLL~\citep{cho2026dsert} & OF & Real & RETL & \na \\
    DTU~\citep{jensen2014dtu} & WB & Real & RD & \na \\
    MVSEC~\citep{zhu2018mvsec} & OF & Real & E & \na \\
    EDS~\citep{hidalgo2022event} & WB & Real & RE & \na \\
    \bottomrule
  \end{tabular}%
  }
\par\vspace{2pt}
\parbox{0.9\linewidth}{\scriptsize OF: optical flow; WB: wide baseline.
Source: Real, Synthetic, or Both. Modalities: R (RGB), E (event),
D (depth), T (thermal), L (projected LiDAR).}
\end{table}
\begin{table*}[t]
    \caption{Cross-modal EPE (px), $>$5px outlier rate, and inference time on one H100 GPU}
    \label{tab:cross_modal_matching}
    \centering
    \begingroup
    \tiny
    \setlength{\tabcolsep}{1.05pt}
    \renewcommand{\arraystretch}{0.82}
    \begin{tabular}{l*{18}{c}c}
      \toprule
      \multirow{4}{*}{Method}
        & \multicolumn{10}{c}{RGB-anchored cross-modal pairs}
        & \multicolumn{8}{c}{Non-RGB cross-modal pairs}
        & \multirow{4}{*}{Inference Time} \\
      \cmidrule(lr){2-11}\cmidrule(lr){12-19}
        & \multicolumn{2}{c}{RGB$\to$Depth}
        & \multicolumn{2}{c}{RGB$\to$Event}
        & \multicolumn{4}{c}{RGB$\to$Thermal}
        & \multicolumn{2}{c}{RGB$\to$LiDAR}
        & \multicolumn{2}{c}{Thermal$\to$Event}
        & \multicolumn{2}{c}{Event$\to$LiDAR}
        & \multicolumn{2}{c}{Thermal$\to$LiDAR}
        & \multicolumn{2}{c}{Depth$\to$Thermal}
        & \\
      \cmidrule(lr){2-3}\cmidrule(lr){4-5}\cmidrule(lr){6-9}\cmidrule(lr){10-11}\cmidrule(lr){12-13}\cmidrule(lr){14-15}\cmidrule(lr){16-17}\cmidrule(lr){18-19}
        & \multicolumn{2}{c}{DTU}
        & \multicolumn{2}{c}{DSERT-RoLL}
        & \multicolumn{2}{c}{MTV}
        & \multicolumn{2}{c}{DSERT-RoLL}
        & \multicolumn{2}{c}{DSERT-RoLL}
        & \multicolumn{2}{c}{DSERT-RoLL}
        & \multicolumn{2}{c}{DSERT-RoLL}
        & \multicolumn{2}{c}{DSERT-RoLL}
        & \multicolumn{2}{c}{MTV}
        & \\
      \cmidrule(lr){2-3}\cmidrule(lr){4-5}\cmidrule(lr){6-7}\cmidrule(lr){8-9}\cmidrule(lr){10-11}\cmidrule(lr){12-13}\cmidrule(lr){14-15}\cmidrule(lr){16-17}\cmidrule(lr){18-19}
        & EPE$\downarrow$ & 5px$\downarrow$
        & EPE$\downarrow$ & 5px$\downarrow$
        & EPE$\downarrow$ & 5px$\downarrow$
        & EPE$\downarrow$ & 5px$\downarrow$
        & EPE$\downarrow$ & 5px$\downarrow$
        & EPE$\downarrow$ & 5px$\downarrow$
        & EPE$\downarrow$ & 5px$\downarrow$
        & EPE$\downarrow$ & 5px$\downarrow$
        & EPE$\downarrow$ & 5px$\downarrow$
        & ms$\downarrow$ \\
      \midrule
      \multicolumn{20}{l}{\textit{Dense matching}} \\
      MINIMA(RoMa)~\cite{ren2025minima} & \underline{36.56} & \underline{67.2} & 13.77 & 62.9 & 8.61 & 15.0 & \textbf{9.57} & \textbf{53.2} & 73.02 & 96.5 & \underline{13.58} & \textbf{63.4} & 57.39 & 95.7 & 63.77 & 96.7 & 24.52 & \textbf{40.2} & 262.4 \\
      MatchAnything(RoMa)~\cite{he2025matchanything} & 37.38 & 77.9 & \underline{11.61} & \underline{44.4} & \underline{3.28} & \underline{14.5} & 11.93 & \underline{55.8} & \underline{52.70} & 92.7 & 15.23 & 65.3 & \underline{30.48} & \underline{91.9} & \underline{42.88} & \underline{94.2} & \underline{23.80} & 46.1 & 206.4 \\
      RoMa~\cite{edstedt2024roma} & 67.72 & 78.5 & 47.00 & 77.4 & 20.06 & 26.6 & 16.44 & 58.3 & 229.23 & 99.9 & 39.68 & 78.6 & 213.78 & 99.7 & 229.67 & 99.8 & 96.98 & 70.3 & 192.3 \\
      RoMa v2~\cite{edstedt2025romav2} & 150.89 & 94.0 & 23.25 & 58.8 & 84.34 & 52.0 & 22.19 & 60.6 & 141.00 & 99.5 & 20.64 & 67.2 & 135.59 & 99.3 & 118.06 & 98.8 & 200.49 & 86.7 & 61.5 \\
      UFM~\cite{zhang2026ufm} & 54.81 & 74.9 & 63.70 & 87.6 & 16.13 & 24.8 & 19.31 & 66.3 & 75.46 & \underline{92.3} & 79.56 & 94.8 & 45.99 & 96.1 & 126.09 & 99.4 & 126.44 & 78.2 & \textbf{27.6} \\
      \ours{} (Ours) & \textbf{12.78} & \textbf{37.5} & \textbf{6.63} & \textbf{38.0} & \textbf{2.48} & \textbf{8.3} & \underline{10.95} & 69.5 & \textbf{8.46} & \textbf{54.3} & \textbf{9.61} & \underline{64.7} & \textbf{9.55} & \textbf{61.0} & \textbf{11.61} & \textbf{78.3} & \textbf{13.45} & \underline{41.2} & \underline{27.8} \\
      \midrule
      \multicolumn{20}{l}{\textit{Sparse matching} (cross-scored on each baseline's match set)} \\
      XoFTR~\cite{tuzcuoglu2024xoftr} & 42.05 & 49.7 & 19.89 & 48.4 & 12.49 & 27.6 & 13.35 & \textbf{47.9} & 109.18 & 93.7 & 34.49 & 66.1 & 125.81 & 95.3 & 118.42 & 98.2 & 36.67 & 50.3 & 21.0 \\
      \quad \ours{} (Ours) & \textbf{4.61} & \textbf{11.6} & \textbf{3.88} & \textbf{18.7} & \textbf{1.90} & \textbf{4.7} & \textbf{6.91} & 56.7 & \textbf{5.70} & \textbf{38.9} & \textbf{6.72} & \textbf{47.8} & \textbf{5.60} & \textbf{43.2} & \textbf{7.06} & \textbf{71.1} & \textbf{5.08} & \textbf{19.0} & 27.7 \\
      MatchAnything (ELoFTR)~\cite{he2025matchanything} & 6.95 & 23.1 & 9.39 & 51.3 & 6.63 & 40.2 & 9.06 & 60.0 & 124.63 & 93.9 & 13.14 & 67.8 & 81.80 & 93.9 & 102.71 & 94.7 & 15.08 & 50.2 & 36.3 \\
      \quad \ours{} (Ours) & \textbf{1.76} & \textbf{2.6} & \textbf{3.77} & \textbf{18.7} & \textbf{1.87} & \textbf{4.4} & \textbf{7.04} & \textbf{57.1} & \textbf{5.82} & \textbf{41.1} & \textbf{6.35} & \textbf{46.9} & \textbf{5.84} & \textbf{42.8} & \textbf{7.45} & \textbf{71.1} & \textbf{2.89} & \textbf{10.2} & 27.8 \\
      MINIMA (LoFTR)~\cite{ren2025minima} & 17.80 & 31.8 & 8.90 & 43.1 & 4.95 & 19.1 & 7.25 & \textbf{43.6} & 52.43 & 84.2 & 11.20 & 57.9 & 65.51 & 87.5 & 39.86 & 91.3 & 9.88 & 24.3 & 21.1 \\
      \quad \ours{} (Ours) & \textbf{2.37} & \textbf{4.8} & \textbf{3.75} & \textbf{19.0} & \textbf{1.79} & \textbf{4.0} & \textbf{6.22} & 52.7 & \textbf{4.55} & \textbf{28.8} & \textbf{6.06} & \textbf{44.1} & \textbf{5.02} & \textbf{34.9} & \textbf{6.35} & \textbf{68.9} & \textbf{3.88} & \textbf{13.1} & 27.6 \\
      \bottomrule
    \end{tabular}%

    \endgroup
  \end{table*}
\begin{table*}[tbp]
\centering
\begingroup
\tiny
\renewcommand{\arraystretch}{0.82}
\begin{minipage}[t]{0.56\textwidth}
    \centering
    \caption{Relative pose estimation benchmark for cross-modal pairs.}
    \label{tab:pose_estimation_final}
    \setlength{\tabcolsep}{1.5pt}
    \begin{tabular}{lccccccc}
    \toprule
    \multirow{2}{*}{Method}
        & \multicolumn{3}{c}{DTU RGB$\to$Depth}
        & \multicolumn{3}{c}{EDS RGB$\to$Event}
        & \multirow{2}{*}{Inference Time} \\
    \cmidrule(lr){2-4}
    \cmidrule(lr){5-7}
        & AUC@5$^\circ\uparrow$ & AUC@10$^\circ\uparrow$ & AUC@20$^\circ\uparrow$
        & AUC@5$^\circ\uparrow$ & AUC@10$^\circ\uparrow$ & AUC@20$^\circ\uparrow$
        & ms$\downarrow$ \\
    \midrule
    MINIMA(RoMa)~\cite{ren2025minima} & \textbf{10.7} & \underline{20.4} & \underline{32.0} & 2.0 & 5.7 & 12.0 & 263.1 \\
    MatchAnything(RoMa)~\cite{he2025matchanything} & 6.5 & 15.3 & 27.0 & \underline{2.9} & \underline{7.2} & \underline{13.9} & 207.9 \\
    RoMa~\cite{edstedt2024roma} & 5.3 & 12.5 & 21.7 & 0.7 & 2.0  & 4.7 & 198.6 \\
    RoMa v2~\cite{edstedt2025romav2} & 1.6 & 3.5 & 5.9 & 1.5 & 4.8 & 9.8 & 61.0 \\
    UFM~\cite{zhang2026ufm} & 6.3 & 14.3 & 25.2 & 0.1 & 0.4 & 1.7 & \textbf{27.7} \\
    \ours{} (Ours) & \underline{10.6} & \textbf{24.7} & \textbf{45.4} & \textbf{8.3} & \textbf{17.8} & \textbf{29.5} & \underline{27.8} \\
    \bottomrule
    \end{tabular}
\end{minipage}
\hfill
\begin{minipage}[t]{0.40\textwidth}
    \centering
    \caption{TartanMatch performance for DSERT-RoLL heavy snow degradation scenes.}
    \label{tab:dsert_degraded_rgb_lidar}
    \setlength{\tabcolsep}{4pt}
    \begin{tabular}{lccccccc}
    \toprule
    \multirow{2}{*}{Scene}
        & \multicolumn{2}{c}{RGB$\to$RGB}
        & \multicolumn{2}{c}{EVENT$\to$EVENT}
        & \multicolumn{2}{c}{EVENT$\to$LIDAR}
        \\
    \cmidrule(lr){2-3}\cmidrule(lr){4-5}\cmidrule(lr){6-7}
        & EPE$\downarrow$ & 2px$\downarrow$ & EPE$\downarrow$ & 2px$\downarrow$ &
        EPE$\downarrow$ & 5px$\downarrow$ \\
    \midrule
    Normal & 2.40 & 26.0 & 2.32 & 28.2 & 9.55 & 61.0\\
    Heavy snow & 30.34 & 99.0 & 0.79 & 3.7 & 3.02 & 14.1\\
    \bottomrule
    \end{tabular}
\end{minipage}
\endgroup
\end{table*}

\begin{figure*}[tbp]
  \centering
  \includegraphics[width=0.91\textwidth]{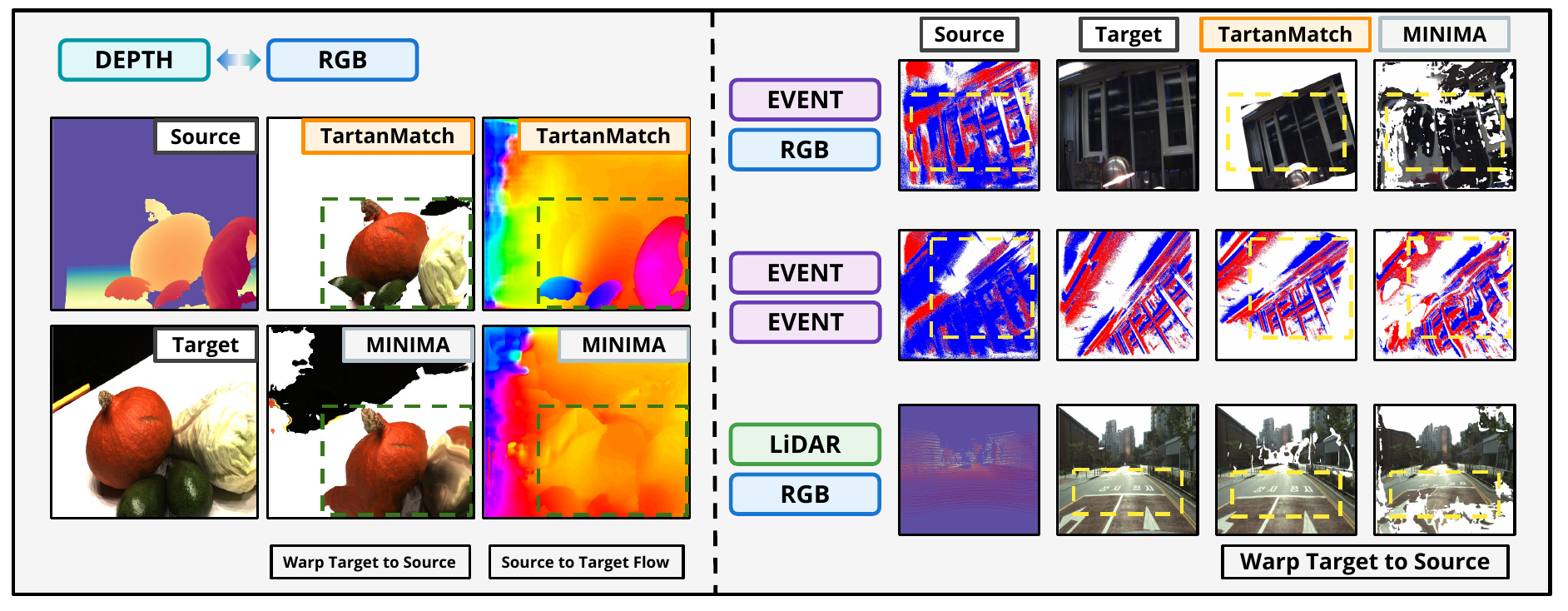}
  \caption{TartanMatch produces more accurate flow and covisibility than MINIMA on various pairs.  }
  \label{fig:qualitative}
\end{figure*}

\begin{table*}[tbp]
\caption{Same-modal EPE (px), $>$2px outlier rate, and inference time on one H100 GPU}
\label{tab:unimodal_matching}
\centering
\begingroup
\tiny
\setlength{\tabcolsep}{1.05pt}
\renewcommand{\arraystretch}{0.82}
    \begin{tabular}{l*{18}{c}c}
    \toprule
    \multirow{3}{*}{Method}
    & \multicolumn{4}{c}{RGB$\to$RGB}
    & \multicolumn{2}{c}{Depth$\to$Depth}
    & \multicolumn{6}{c}{Event$\to$Event}
    & \multicolumn{4}{c}{Thermal$\to$Thermal}
    & \multicolumn{2}{c}{LiDAR$\to$LiDAR}
    & \multirow{3}{*}{Inference Time} \\
    \cmidrule(lr){2-5}\cmidrule(lr){6-7}\cmidrule(lr){8-13}\cmidrule(lr){14-17}\cmidrule(lr){18-19}
    & \multicolumn{2}{c}{DTU} & \multicolumn{2}{c}{DSERT-RoLL}
    & \multicolumn{2}{c}{DTU}
    & \multicolumn{2}{c}{MVSEC 20Hz} & \multicolumn{2}{c}{MVSEC 45Hz} & \multicolumn{2}{c}{DSERT-RoLL}
    & \multicolumn{2}{c}{MTV} & \multicolumn{2}{c}{DSERT-RoLL}
    & \multicolumn{2}{c}{DSERT-RoLL}
    & \\
    \cmidrule(lr){2-3}\cmidrule(lr){4-5}\cmidrule(lr){6-7}\cmidrule(lr){8-9}\cmidrule(lr){10-11}\cmidrule(lr){12-13}\cmidrule(lr){14-15}\cmidrule(lr){16-17}\cmidrule(lr){18-19}
    & EPE$\downarrow$ & 2px$\downarrow$ & EPE$\downarrow$ & 2px$\downarrow$
    & EPE$\downarrow$ & 2px$\downarrow$
    & EPE$\downarrow$ & 2px$\downarrow$ & EPE$\downarrow$ & 2px$\downarrow$ & EPE$\downarrow$ & 2px$\downarrow$
    & EPE$\downarrow$ & 2px$\downarrow$ & EPE$\downarrow$ & 2px$\downarrow$
    & EPE$\downarrow$ & 2px$\downarrow$
    & ms$\downarrow$ \\
    \midrule
    \multicolumn{20}{l}{\textit{Dense matching}} \\
    MINIMA(RoMa)~\cite{ren2025minima} & 12.28 & 30.9 & 4.09 & 29.6 & 29.86 & 73.9 & 2.06 & 32.6 & 1.92 & 30.4 & 4.80 & 44.8 & 1.07 & 4.3 & 6.52 & 43.3 & 13.21 & 81.8 & 260.2 \\
    MatchAnything(RoMa)~\cite{he2025matchanything} & 12.49 & 32.3 & 3.95 & 27.4 & \underline{26.54} & 71.9 & 1.84 & 31.4 & 1.45 & 23.2 & 4.80 & 44.4 & \textbf{0.56} & \textbf{3.6} & 6.83 & 43.7 & 13.78 & 82.1 & 206.9 \\
    RoMa~\cite{edstedt2024roma} & 11.52 & \underline{29.5} & 4.04 & 28.0 & 37.26 & 80.7 & \underline{1.33} & \underline{21.7} & 0.96 & 10.4 & 6.19 & 47.2 & 1.32 & 4.6 & 6.96 & 43.7 & \underline{13.01} & \underline{78.4} & 191.3 \\
    RoMa v2~\cite{edstedt2025romav2} & \textbf{4.68} & \textbf{20.3} & \underline{2.55} & \textbf{24.2} & \underline{26.54} & \underline{63.8} & 1.46 & 25.3 & 0.88 & 9.4 & 3.45 & 42.6 & \underline{0.69} & \underline{3.9} & \underline{3.21} & \textbf{33.3} & 13.92 & 82.0 & 60.6 \\
    UFM~\cite{zhang2026ufm} & \underline{5.33} & 32.0 & 2.75 & 26.8 & 37.32 & 81.8 & 1.68 & 30.1 & \underline{0.77} & \underline{7.3} & 10.68 & 86.0 & 0.70 & 5.4 & 4.50 & 37.6 & 14.57 & 84.1 & 27.5 \\
    E-RAFT~\cite{gehrig2021eraft} & \na & \na & \na & \na & \na & \na & 1.69 & 30.1 & 0.83 & 7.7 & \underline{2.40} & 29.3 & \na & \na & \na & \na & \na & \na & \textbf{19.1} \\
    EMatch~\cite{zhang2024ematch} & \na & \na & \na & \na & \na & \na & 1.86 & 34.1 & 0.92 & 10.5 & 2.44 & \textbf{27.6} & \na & \na & \na & \na & \na & \na & \underline{24.7} \\
    \ours{} (Ours) & 5.74 & 36.8 & \textbf{2.40} & \underline{26.0} & \textbf{5.67} & \textbf{45.4} & \textbf{1.17} & \textbf{18.6} & \textbf{0.70} & \textbf{5.6} & \textbf{2.32} & \underline{28.2} & 0.77 & 6.2 & \textbf{3.07} & \underline{34.3} & \textbf{5.89} & \textbf{48.6} & 27.8 \\
    \midrule
    \multicolumn{20}{l}{\textit{Sparse matching} (cross-scored on each baseline's match set)} \\
    XoFTR~\cite{tuzcuoglu2024xoftr} & 15.57 & 33.9 & 3.09 & 25.0 & 25.68 & 67.6 & 4.27 & 36.8 & 5.45 & 39.2 & 7.95 & 62.3 & 4.51 & 8.2 & 6.17 & 45.6 & 10.28 & 80.0 & 24.8 \\
    \quad \ours{} (Ours) & \textbf{2.33} & \textbf{19.9} & \textbf{2.06} & \textbf{21.8} & \textbf{3.25} & \textbf{25.4} & \textbf{0.81} & \textbf{8.8} & \textbf{0.51} & \textbf{1.9} & \textbf{1.92} & \textbf{21.8} & \textbf{0.70} & \textbf{5.0} & \textbf{2.75} & \textbf{29.9} & \textbf{3.20} & \textbf{33.0} & 27.6 \\
    MatchAnything (ELoFTR)~\cite{he2025matchanything} & 6.33 & 31.7 & 2.80 & 25.3 & 6.92 & 65.4 & 1.98 & 34.8 & 1.82 & 31.3 & 6.39 & 65.1 & 1.12 & 11.2 & 7.39 & 52.9 & 10.07 & 85.0 & 37.4 \\
    \quad \ours{} (Ours) & \textbf{1.90} & \textbf{18.9} & \textbf{1.86} & \textbf{20.0} & \textbf{1.44} & \textbf{13.6} & \textbf{0.75} & \textbf{7.2} & \textbf{0.48} & \textbf{1.4} & \textbf{1.96} & \textbf{22.9} & \textbf{0.73} & \textbf{5.7} & \textbf{2.76} & \textbf{30.1} & \textbf{4.17} & \textbf{38.0} & 27.7 \\
    MINIMA (LoFTR)~\cite{ren2025minima} & 9.88 & 38.2 & 2.85 & 29.7 & 14.36 & 63.8 & 1.44 & 19.2 & 1.30 & 14.2 & 5.38 & 56.7 & 1.46 & 12.5 & 5.76 & 49.2 & 4.68 & 61.4 & 22.3 \\
    \quad \ours{} (Ours) & \textbf{1.84} & \textbf{15.2} & \textbf{1.82} & \textbf{19.1} & \textbf{1.72} & \textbf{15.6} & \textbf{0.66} & \textbf{4.6} & \textbf{0.45} & \textbf{0.9} & \textbf{1.82} & \textbf{20.7} & \textbf{0.70} & \textbf{5.0} & \textbf{2.71} & \textbf{28.3} & \textbf{2.14} & \textbf{24.5} & 27.7 \\
    \bottomrule
    \end{tabular}%
\endgroup
\end{table*}

\section{Experiments}
\label{sec:experiments}

\subsection{Implementation Details}
\label{sec:implementation}

\paragraph{Training setup.}
We train TartanMatch in two stages (Sec.~\ref{sec:training_procedure}) at $560$ input resolution on $8$ H100 GPUs, using AdamW ($\beta=(0.9, 0.95)$, weight decay $0.05$) with a cosine schedule, 10\% linear warmup, gradient-norm clipping at $1.0$, bf16 mixed precision, and an effective batch size of $48$, each batch is symmetrized following UFM. Each epoch draws 650k modality-pair samples stratified over the 25 modality pairs.
Stage 1 trains for $10$ epochs with learning rates $10^{-4}$ for the input projections and $10^{-5}$ for the information-sharing transformer and DPT heads; Stage 2 fine-tunes end-to-end for $35$ epochs, adding the encoder at $5\times10^{-7}$ with the remaining rates unchanged.
The loss of Sec.~\ref{sec:shared_backbone} uses $\alpha=0.5$ and $c=0.24$.

\paragraph{Preprocessing}
For RGB inputs, we follow UFM's augmentation pipeline.
For simulated events, the integration window between the two frames is randomly clipped to $[10\,\text{ms}, 100\,\text{ms}]$ to diversify event sparsity. We inject random noise events drawn i.i.d. uniformly over the sensor plane and time window with random polarity at a per-sample rate of $U(0, 0.5)$ events/pixel before voxelization. For raw 16-bit radiometric thermal data like FIReStereo, we convert each frame to an 8-bit image by per-image normalization (a 1--99th-percentile clip followed by min--max scaling) and then apply CLAHE and bilateral filtering~\cite{dhrafani2025firestereo}. For depth and LiDAR, we apply $\log(1+x)$ to valid values, normalize them per sample using the 2.5--97.5th percentiles, and concatenate a binary validity mask to form the two-channel representation.

\subsection{Evaluation protocol}
We use five real-world datasets for zero-shot benchmark: MTV~\cite{liu2023mtv}, DSERT-RoLL~\cite{cho2026dsert}, DTU~\cite{jensen2014dtu}, MVSEC~\cite{zhu2018mvsec}, and EDS~\cite{hidalgo2022event}. Refer to Table~\ref{tab:datasets} for the modality composition of each dataset. On MTV, we use the race-track, country-road, and park scenes. Similarly, we select an Outdoor Day1 sequence for MVSEC, following E-RAFT. On DSERT-RoLL, we evaluate same-modality and cross-modality matching on the normal scenes and degradation robustness on the heavy-snow scene. For EDS, we sample pairs with pose differences greater than $1^\circ$, $5^\circ$, $10^\circ$, and $20^\circ$.

For each modality pair, we compare against the strongest available baseline with publicly released weights, ranging from pair-specific models to cross-modal correspondence models.
UFM~\cite{zhang2026ufm} is TartanMatch's RGB-pretrained starting point and therefore measures the benefit of multi-modal adaptation; MINIMA~\cite{ren2025minima} and MatchAnything~\cite{he2025matchanything} are the strongest prior methods targeting broad cross-modal correspondence. RoMa~\cite{edstedt2024roma} provides the RGB-only counterpart to the RoMa-based variants of MINIMA and MatchAnything, whereas RoMa~v2~\cite{edstedt2025romav2} serves as a state-of-the-art RGB-only baseline. Following MINIMA, we evaluate RGB-only baselines on non-RGB modalities by rendering each sensor measurement to a three-channel image with a fixed colormap (RGB-like rendering). For event correspondence baselines, we include E-RAFT~\cite{gehrig2021eraft} and EMatch~\cite{zhang2024ematch}, both are specialized in event and trained on DSEC. For sparse correspondence, we additionally evaluate XoFTR~\cite{tuzcuoglu2024xoftr} and the LoFTR/ELoFTR variants of MINIMA and MatchAnything. For comparisons with sparse matching models, we score each baseline at the source pixels it selects and sample TartanMatch's flow predictions at the same locations.

Inference uses each model's internal resolution.
We report Endpoint Error (EPE) ($\downarrow$), the outlier ratio above 2px for same-modality matching and above 5px for cross-modality matching ($\downarrow$), and Area Under Curve (AUC) ($\uparrow$) for relative pose estimation. We evaluate all models on single H100 GPU and report their average inference time per sample. The best and second-best results for each benchmark are shown in bold and underlined, respectively.

\subsection{Cross-Modal Dense Correspondence}
\label{sec:cross_modal}

Table~\ref{tab:cross_modal_matching} reports zero-shot performance on RGB-anchored (one view RGB) and non-RGB cross-modal (neither view RGB) pairs against dense and sparse matching baselines.
Among RGB-anchored pairs, TartanMatch leads on most settings, reducing EPE over the strongest baseline by 65\% on RGB$\to$Depth and 43\% on RGB$\to$Event.
The gains are larger for projected-LiDAR pairs, which no baseline sees during training: the best competitor records at least 30.48px EPE, whereas TartanMatch remains below 12px.
On Event$\to$LiDAR and Thermal$\to$LiDAR, TartanMatch reduces EPE by 3.2$\times$ and 3.7$\times$ and lowers the 5px outlier ratio by 31 and 16 points, respectively.
TartanMatch also consistently outperforms sparse matching baselines when evaluated only at their selected correspondences.
Fig.~\ref{fig:qualitative} further shows more complete warps and coherent flow than MINIMA across representative same-modal and cross-modal pairs.

Table~\ref{tab:pose_estimation_final} evaluates pose estimation on DTU and EDS for RGB$\rightarrow$depth and RGB$\rightarrow$event, respectively.
On DTU, TartanMatch raises AUC@20$^\circ$ by 42\% over the strongest baseline while matching it at AUC@5$^\circ$.
On EDS, where all baselines struggle, it improves AUC across thresholds by 2.1--2.9$\times$ while running 7.5$\times$ faster than MatchAnything.
Together, these results demonstrate strong cross-modal correspondence and  pose estimation performance.

\subsection{Same-Modality Dense Correspondence}
\label{sec:within_modal}
Table~\ref{tab:unimodal_matching} reports same-modal zero-shot performance for all five modalities against the strongest baseline available.
The largest gains appear on the modalities that RGB matchers transfer worst: TartanMatch reduces EPE by 4.7$\times$ on Depth$\to$Depth and by 55\% on LiDAR$\to$LiDAR.
On Event$\to$Event, TartanMatch outperforms state-of-the-art modality specialists, E-RAFT and EMatch, the dedicated event flow estimation models.
On RGB and thermal pairs, TartanMatch stays within 1.1px and 0.3px of the best baseline.
Similarly to cross-modal case, TartanMatch outperforms all sparse matching methods in same-modal matching.
Overall, modality-specific models generalize poorly beyond their own modality, whereas TartanMatch transfers across all five modalities and even surpasses specialists on some modalities.

\subsection{Degraded Scenarios}
\label{sec:degraded_scenes}
TartanMatch's any-pair matching capability enables robust correspondence estimation when individual sensors degrade.
Table~\ref{tab:dsert_degraded_rgb_lidar} evaluates TartanMatch on the heavy-snow sequences of DSERT-RoLL, testing robustness under RGB degradation.
Snow renders RGB$\to$RGB matching unusable: EPE grows from 2.40 to 30.34 and the 2px outlier ratio reaches 99\%.
The event-based pairs stay accurate under the same conditions, with 0.79 EPE on Event$\to$Event and 3.02 EPE on Event$\to$LiDAR.
An any-pair mode lets a platform shift correspondence to sensors that stay informative.

\begin{table}[tbp]
    \caption{Ablation studies. (a) trains on all event
   datasets; (b) trains on all depth datasets;
   (c) uses the full training mixture.}
     \label{tab:ablations}
     \centering
     \small
     \resizebox{0.9\columnwidth}{!}
     {%
     \setlength{\tabcolsep}{5pt}
     \begin{tabular}{@{}lcrr@{}}
     \toprule
     & & \multicolumn{2}{c}{EPE $\downarrow$} \\
     \cmidrule(l){3-4}
     Setting & Test set & Cross-modality & Same-modality \\
     \midrule
     \multicolumn{4}{@{}l}{\textbf{(a) Event representation}} \\
     \quad Polarity-accumulated color image & \multirow{2}{*}{\itshape DSERT-RoLL} & 10.56 & 3.44 \\
     \quad Voxel volume (Ours)              &  & \textbf{7.46} & \textbf{2.46} \\
     \addlinespace
     \multicolumn{4}{@{}l}{\textbf{(b) Depth representation}} \\
     \quad Color-mapped image      & \multirow{3}{*}{\itshape DTU} & 12.30 & 6.30 \\
     \quad Log-normalized          &  & 11.49 & 6.26 \\
     \quad $+$ validity mask (Ours)&  & \textbf{11.33} & \textbf{5.64} \\
     \addlinespace
     \multicolumn{4}{@{}l}{\textbf{(c) Training schedule}} \\
     \multirow{2}{*}{\quad One-stage (joint)}
       & \itshape DSERT-RoLL & 10.21 & 3.46 \\
       & \itshape DTU        & 14.01 & 6.95 \\
     \multirow{2}{*}{\quad Two-stage (Ours)}
       & \itshape DSERT-RoLL & \textbf{9.80} & \textbf{3.42} \\
       & \itshape DTU        & \textbf{12.38} & \textbf{5.71} \\
     \bottomrule
   \end{tabular}
     }
   \end{table}

\begin{table}[tbp]
\caption{EPE on the TartanAirV2. Single-modality-pair models use their best checkpoint per pair; the all-modality model is a single checkpoint. $\Delta$ is the relative improvement of all-modality over single-pair.}
\label{tab:modality_transfer_val}
\centering
\resizebox{0.6\linewidth}{!}{%
\begin{tabular}{lrrr}
\toprule
Pair & Single-pair & All-mod. & $\Delta$ (\%) \\
\midrule
\multicolumn{4}{l}{\emph{Same-modality}} \\
\quad R$\rightarrow$R & \textbf{0.82} & 1.00 & $-22.0$ \\
\quad D$\rightarrow$D & \textbf{0.73} & 0.90 & $-23.3$ \\
\quad E$\rightarrow$E & 1.29 & \textbf{1.22} & $+5.4$ \\
\quad T$\rightarrow$T & \textbf{3.14} & 3.24 & $-3.2$ \\
\quad L$\rightarrow$L & 7.84 & \textbf{5.00} & $+36.2$ \\
\cmidrule(lr){1-4}
\quad \emph{Average (5)} & \emph{2.76} & \textbf{2.27} & $+17.8$ \\
\midrule
\multicolumn{4}{l}{\emph{Cross-modality}} \\
\quad R$\rightarrow$D & 3.16 & \textbf{2.89} & $+8.5$ \\
\quad R$\rightarrow$E & 2.66 & \textbf{1.69} & $+36.5$ \\
\quad R$\rightarrow$T & \textbf{2.28} & 2.34 & $-2.6$ \\
\quad R$\rightarrow$L & 8.78 & \textbf{6.01} & $+31.6$ \\
\quad D$\rightarrow$E & 3.12 & \textbf{1.85} & $+40.7$ \\
\quad D$\rightarrow$T & 5.93 & \textbf{5.04} & $+15.0$ \\
\quad D$\rightarrow$L & 4.85 & \textbf{3.19} & $+34.2$ \\
\quad E$\rightarrow$T & 5.29 & \textbf{3.85} & $+27.2$ \\
\quad E$\rightarrow$L & 14.20 & \textbf{5.47} & $+61.5$ \\
\quad T$\rightarrow$L & 13.02 & \textbf{9.37} & $+28.0$ \\
\cmidrule(lr){1-4}
\quad \emph{Average (10)} & \emph{6.33} & \textbf{4.17} & $+34.1$ \\
\bottomrule
\end{tabular}%
}
\par\vspace{2pt}
\parbox{0.9\linewidth}{\scriptsize\centering R (RGB), E (event),
D (depth), T (thermal), L (projected LiDAR).}
\end{table}

\subsection{Ablations}
\label{sec:ablations}
Our ablation investigates three questions:
(i) whether the proposed sensor-specific input representations outperform fixed RGB-like renderings;
(ii) whether the two-stage training strategy improves correspondence estimation over one-stage end-to-end training; and
(iii) whether jointly training on multiple modality pairs benefits the performance of each individual pair.
Owing to compute and time constraints, the ablation variants follow TartanMatch's hyperparameters described in Section~\ref{sec:implementation}, but train at smaller scale, specified in each paragraph below. For ablation  \textbf{(a)} and \textbf{(c)}, we train the model for 20 epochs, which we find sufficient for convergence. For the ablation \textbf{(b)}, we use the full training mixture and recipe.
We report mean EPE for cross-modality and same-modality matching.

\textbf{(a) Input representation.} Each variant trains on pairs from $\{RGB,Event\}$ and $\{RGB,Depth\}$ using every training dataset that provides such modality pairs. (See Table~\ref{tab:datasets})
We evaluate the event variants on DSERT-RoLL, the only benchmark that pairs event with non-event modalities and thus measures both cross and same-modal performance, and likewise the depth variants on DTU.
In Table~\ref{tab:ablations}~(a), the event voxel volume reduces cross-modal (RGB$\leftrightarrow$Event)
EPE from 10.56 to 7.46 over the fixed RGB-like rendering using event polarity accumulation.
In Table~\ref{tab:ablations}~(b), log-normalized depth with a validity mask improves over the color-mapped image on both cross-modal (RGB$\leftrightarrow$Depth) and same-modal (Depth$\to$Depth) matching, 12.30 to 11.33 and 6.30 to 5.64, respectively. These results validate the effectiveness of our modality-specific input adaptation, showing that simply rendering non-RGB modalities into RGB-like colormap images leads to suboptimal correspondence estimation.

\textbf{(b) One-stage vs.\ two-stage training.} This ablation follows the full TartanMatch setup: we compare the two-stage recipe against training the full model end to end for 45 epochs on the complete training mixture.
Table~\ref{tab:ablations}~(c) shows consistent gains: cross-modal EPE decreases from 10.21 to 9.80 on DSERT-RoLL and from 14.01 to 12.38 on DTU, with same-modal EPE also improving. These results show that the two-stage schedule enables more effective multi-modal adaptation than one-stage training, improving both cross-modal and same-modal matching.

\textbf{(c) Does multi-modal training help?} To limit training cost without excluding any modality combination, we train 15 pair-specific models, one for each unordered pair of the five modalities (5 same-modal and 10 cross-modal), and compare them with a single model jointly trained on all 25 ordered pairs in TartanAirV2. We evaluate them on the validation split of TartanAirV2. As shown in Table~\ref{tab:modality_transfer_val}, joint training reduces mean EPE by 34.1\% for cross-modal pairs and 17.8\% for same-modal pairs, improving 11 of the 15 evaluated pairs. The largest gain is 61.5\% on Event$\to$LiDAR, which also has the highest pair-specific EPE. All four regressions are at most 0.2\,px. We hypothesize that joint multi-modal training favors modality-invariant semantic cues over modality-specific low-level cues, benefiting challenging pairs at a small cost in fine-grained matching on those with sub-pixel accuracy. These results demonstrate strong positive transfer where pair-specific training struggles, with minimal cost on the remaining pairs, supporting unified multi-modal training.

\section{Conclusion}
\label{sec:conclusion}

We presented TartanMatch, a unified dense correspondence model that estimates dense flow and covisibility for any pair between RGB, event, thermal, depth, and projected LiDAR. A single backbone, inherited from a pretrained RGB correspondence model, serves all five modalities; the only per-modality parameters are a thin input projection over a representation chosen for each sensor. Trained on a data suite spanning 25 modality pairs, TartanMatch achieves state-of-the-art zero-shot accuracy on same-modality and cross-modality pairs, providing the first dense any-pair baseline.
Because each modality costs only an input representation and a thin projection, future work can extend TartanMatch to new sensors, such as radar, with few additional parameters. We further plan to collect real data for underrepresented modalities, such as thermal and LiDAR, and to apply any-pair correspondence to downstream tasks such as cross-modal odometry, cross-sensor calibration, and place recognition.

\section*{Acknowledgments}
This work was supported by the DSTA under contract number \#DST000EC124000205.
Computational resources for this research were provided by Bridges-2 at the Pittsburgh Supercomputing Center under ACCESS allocation cis220039p.
The ACCESS program is supported by NSF awards 2138259, 2138286, 2138307, 2137603, and 2138296.

\setlength{\bibsep}{0pt plus 0.3pt}
\renewcommand*{\bibfont}{\fontsize{7.5}{8.5}\selectfont}

\end{document}